\documentclass{article}

\usepackage[preprint]{neurips_2023}

\usepackage{wrapfig}
\usepackage[utf8]{inputenc} 
\usepackage[T1]{fontenc}    
\usepackage{hyperref}       
\usepackage{url}            
\usepackage{booktabs}       
\usepackage{amsfonts}       
\usepackage{nicefrac}       
\usepackage{microtype}      
\usepackage{xcolor}         
\usepackage{graphicx}
\usepackage{multirow}
\usepackage{comment}
\usepackage{subfigure}
\usepackage{amsmath}
\usepackage{enumitem}
\usepackage{wrapfig}
\usepackage{bbding}
\usepackage{float}

\title{Generative Text-Guided 3D Vision-Language Pretraining for Unified Medical Image Segmentation}

\author{
Yinda Chen$^{1,2,*}$, 
Che Liu$^{3,}$\thanks{$\*$ Equal Contribution.}, 
Wei Huang$^{1}$, 
Sibo Cheng$^{3}$,
Rossella Arcucci$^{3}$,
Zhiwei Xiong$^{1,2}$
\\
$^{1}$University of Science and Technology of China \quad \\
$^{2}$Institute of Artificial Intelligence, Hefei Comprehensive National Science Center \quad \\
$^{3}$Imperial College London \\
{\tt\small\{cyd0806, weih527@\}mail.ustc.edu.cn }\\
{\tt\small\{lche.liu21,sibo.cheng, r.arcucci@\}imperial.ac.uk, }\\
{\tt\small\ zwxiong@ustc.edu.cn}\\
}
\begin{document}

\maketitle
\begin{abstract}
Vision-Language Pretraining (VLP) has demonstrated remarkable capabilities in learning visual representations from textual descriptions of images without annotations.
Yet, effective VLP demands large-scale image-text pairs, a resource that suffers scarcity in the medical domain. 
Moreover, conventional VLP is limited to 2D images while medical images encompass diverse modalities, often in 3D, making the learning process more challenging.
To address these challenges, we present \textbf{G}enerative \textbf{T}ext-\textbf{G}uided 3D Vision-Language Pretraining for Unified \textbf{M}edical Image Segmentation (GTGM), a framework that extends of VLP to 3D medical images without relying on paired textual descriptions.
Specifically, GTGM utilizes large language models (LLM) to generate medical-style text from 3D medical images. This synthetic text is then used to supervise 3D visual representation learning.
Furthermore, a negative-free contrastive learning objective strategy is introduced  to cultivate consistent visual representations between augmented 3D medical image patches, which effectively mitigates the biases associated with strict positive-negative sample pairings.
We evaluate GTGM on three imaging modalities - Computed Tomography (CT), Magnetic Resonance Imaging (MRI), and electron microscopy (EM) over 13 datasets.
GTGM's superior performance across various medical image segmentation tasks underscores its effectiveness and versatility, by enabling VLP extension into 3D medical imagery while bypassing the need for paired text.
\end{abstract}

\section{Introduction}
\label{sec: intro}
Vision-Language Pretraining (VLP) has achieved significant progress in~\cite{clip,flip,aflip,alayrac2022flamingo}, owing to its capabilities in learning visual representations from textual descriptions of images without annotations. 
While VLP has been introduced to 2D medical image analysis recently, existing medical VLP works rely heavily on textual descriptions written by experienced experts, and the domain of 3D medical VLP remains largely unexplored~\cite{convirt,huang2021gloria,boecking2022making,mgca,zhou2023advancing}.
Despite the fact that 3D medical images, such as Computed Tomography (CT), Magnetic Resonance Imaging (MRI), and electron microscopy (EM), typically contain more valuable and clinically relevant information compared to 2D images, their utilization is hindered primarily due to the lack of associated 3D medical image-text datasets.
Additionally, certain modalities of medical imaging, like EM, often do not have corresponding textual descriptions in real-world applications. 

To address the above challenges, we propose a framework called \textbf{G}enerative \textbf{T}ext-\textbf{G}uided 3D Vision-Language Pretraining for Unified \textbf{M}edical Image Segmentation (GTGM).
GTGM leverages the power of VLP in 3D medical image analysis by employing large language models (LLM) to generate medically relevant textual descriptions for 3D medical images.
The main goal of GTGM is to learn general and robust 3D representations from these synthetic textual descriptions rather than specific organs and modalities. The differences between GTGM and existing image self-supervised algorithms can be summarized in Figure \ref{fig:framework}. GTGM integrates two learning objectives: acquiring visual-textual invariants from 3D medical images and synthetic text and extracting visual invariants from augmented 3D medical images.
To learn general 3D visual representations, we introduce a negative-free contrastive learning strategy. This strategy aims to disentangle the variables in the latent space, rather than following traditional contrastive learning, which may carry biases due to the stringent assumption of one-to-one positive sample pairings.

We evaluate GTGM across various medical image segmentation tasks, covering commonly used modalities like CT and MRI over 10 datasets. 
We also extend our evaluation to the challenging modality of EM, especially the neuron segmentation task, over three datasets and multiple species.
It is noteworthy that the inherent challenges of EM tasks, such as complex neuronal structures, inconsistent image quality, dense packing, and structural heterogeneity, make them significantly more difficult than other modalities \cite{liu2022biological,huang2022semi}.
Impressively, our GTGM attains state-of-the-art (SOTA) results on all EM tasks and nearly all CT and MRI tasks, despite not utilizing real textual descriptions during VLP. This accomplishment underscores the efficacy of synthetic text in guiding 3D medical VLP, which indicates the adaptability and potential of GTGM across a broad spectrum of 3D medical VLP applications.
The contributions of this paper are as follows:
\begin{figure}[tb]
    \centering
    \includegraphics[width = \linewidth]{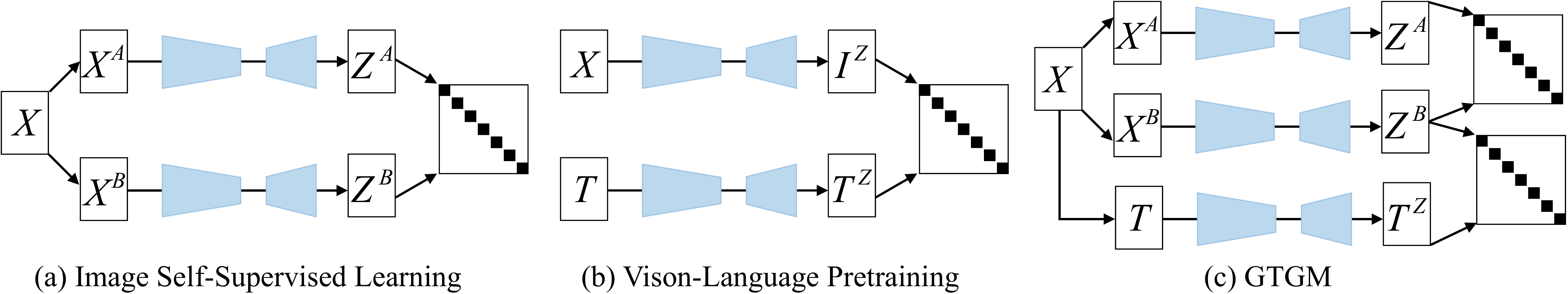}
    \caption{Comparison between our proposed approach and mainstream self-supervised learning (SSL) methods, where $X$ and $T$ represent images and text, respectively.  (a)Image-only SSL with augmented views. 
    (b) Pretraining with paired images and corresponding text.
    (c) Our GTGM framework pretrained with synthetic text-guided VLP and augmented-guided SSL.}
    \label{fig:framework}

\end{figure}
\begin{itemize}[leftmargin=*]
    \item Our GTGM framework is the first to showcase the effectiveness of 3D medical VLP, capable of learning 3D visual representations independent of specific organs or modalities.
    This fills the void in 3D medical visual learning within the scope of VLP.
    Furthermore, GTGM's ability to pretrain without the need for expert-generated real text significantly alleviates one of the major challenges in medical VLP: the lack of large-scale image-text pairs.
    \item GTGM demonstrates superior performance and versatility across various medical image segmentation tasks, supporting different modalities like MRI, CT, and EM, and is adaptable to different species such as human and \textit{Drosophila}. 
    Moreover, GTGM excels in segmenting extremely small and densely packed structures in EM neuron images, expanding its applicability beyond organ and lesion segmentation in CT and MRI images.
    \item GTGM's performance across diverse modalities, organs, and species, as well as its ability to handle varying densities and sizes of segmented objects, indicates its proficiency in learning a comprehensive and robust 3D medical image representation. 
    In other words, GTGM provides an opportunity to generalize novel tasks through text-driven zero-shot medical image segmentation.
\end{itemize}

\section{Related Work}
\label{sec: related}
\noindent\textbf{Image Self-supervised Learning\hspace{2mm}}
Self-supervised learning (SSL) has made significant advancements in computer vision by leveraging pretraining tasks without the need for annotations, as demonstrated by various pretext tasks~\cite{doersch2015unsupervised,gidaris2018unsupervised,noroozi2016unsupervised,zhang2016colorful}.  Recently, contrastive learning has emerged as the standard method in SSL~\cite{byol,barlowtwins,pirl,simsiam,vicreg}.
To address the limitations of traditional contrastive learning, such as the requirement for large batch sizes and strong augmentations~\cite{moco,simclr}, BYOL and BarlowTwins~\cite{byol,barlowtwins} employ a dual-branch structure to align the embeddings of two augmented images, eliminating the need for negative samples in contrastive learning. SimSiam~\cite{simsiam} demonstrates the importance of the stop-gradient mechanism on the dual-encoder, introducing a model without negative samples.
In the context of SSL tailored for medical images, PCRLv2~\cite{pcrlv2} combines contrastive learning with reconstruction pretasks. However, PCRLv2 has limitations in generalization across modalities, particularly in the case of extremely dense and small structures in EM images.

\noindent\textbf{Medical Vision-Language Pretraining\hspace{2mm}}
Medical VLP~\cite{convirt} has been introduced to integrate textual information into medical image SSL. However, the exploration of medical SSL VLP is primarily limited to 2D images, mainly due to the intricacy of medical reports and the scarcity of large-scale medical image-text datasets. 
Nonetheless, in the medical domain, 3D medical images (such as MRI, CT, and EM) assume a vital role and offer richer and more valuable information compared to their 2D medical images.
Studies such as~\cite{convirt,huang2021gloria,mgca,chexzero} concentrate on the chest X-ray (CXR) domain; however, their applicability to other medical image modalities, including MRI, CT, and various 3D medical images, is yet to be established.
In their work~\cite{liu2023clip}, the authors develop a CT segmentation method that incorporates manually generated text describing the organs present in the image, based on corresponding annotations. However, their approach is limited by full supervision and the scale of annotations. Furthermore, their method can only process CT images. In recent works~\cite{butoi2023universeg,ye2023uniseg}, methods are proposed that can process 3D medical segmentation tasks with different modalities. However, these approaches require large-scale well-annotated 3D medical images for supervised pretraining. 
Moreover, UniSeg~\cite{ye2023uniseg} lacks the ability to learn rich textual information as their manually designed prompts only indicate the type of task without describing the images.
Despite its importance, the generalizability of VLP to a wider range of medical applications is limited by the absence of publicly available datasets containing 3D medical image-text pairs, as well as the inability of experienced experts to describe certain modalities such as EM images.

\section{Method}
\label{sec: method}
\subsection{Overview}
\begin{figure}[htb]
    \centering
    \includegraphics[width = \linewidth]{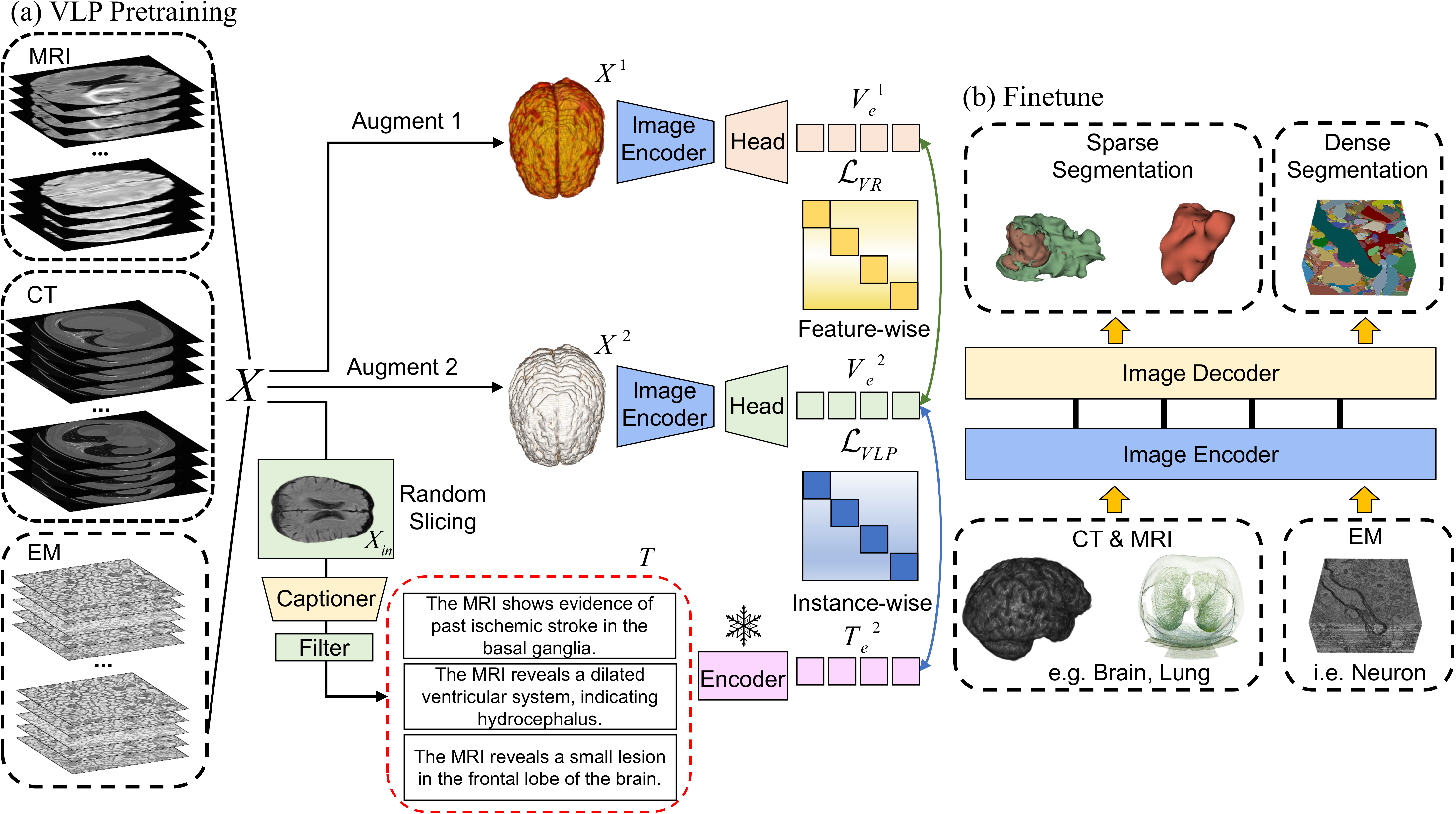}
    \caption{(a) The pipeline of GTGM, where we perform random cropping on medical images and utilize a pretrained generator to generate corresponding text. GTGM parallel learns visual invariants (feature-wise) and visual-textual invariants (instance-wise). (b) The finetuning process, where we employ the image encoder weights obtained during pretraining, along with a limited amount of labeled data, to perform downstream medical image segmentation tasks, including CT, MRI, and EM modalities. The \SnowflakeChevron\hspace{0.23mm} represents frozen weights.}
    \label{fig:main}
 
\end{figure}

Our GTGM model is designed to learn general representations of unannotated 3D medical images from synthetic textual descriptions. Like other VLP models, GTGM incorporates both a visual encoder $f_I(\cdot)$ and a text encoder $f_T(\cdot)$ to extract representations from images and text respectively. However, GTGM uniquely leverages synthetic textual descriptions rather than the real paired text of 3D medical images, given the lack of public 3D medical image-text datasets. The framework is depicted in Figure \ref{fig:main}.

\subsection{Generating Textual Descriptions}
In the process of generating textual descriptions for medical images, we designate a generator $G(\Theta)$, initialize with BLIP~\cite{li2022blip} pretrained weight. This generator is subsequently finetuned on the MedICAT~\cite{subramanian2020medicat} dataset, endowing the synthetic textual descriptions with biomedical style. It is crucial to underscore that MedICAT~\cite{subramanian2020medicat} solely comprises 2D images drawn from biomedical literature, with the associated captions serving as textual descriptions rather than the real description from clinical expertise. Consequently, the generation phase is not limited to any real radiology datasets. During the finetuning phase, we consider a 2D medical image $I$ and its corresponding textual description $T=\{t_1,t_2,...,t_n\}$ from MedICAT~\cite{subramanian2020medicat}, where $n$ is indicative of the textual description's length. The primary objective here is to amplify the conditional probability of the text given one image $I$:
\begin{equation}
P(T \mid I)=\prod_{i=1}^n P\left(t_i \mid I, t_{<i}\right),
\end{equation}
where $t_{<i} = \{t_1, t_2, \ldots, t_{i-1}\}$ denotes the generated text tokens. The conditional probability of each token $t_i$ can be computed as:
\begin{equation}
P\left(t_i \mid I, t_{<i}\right)={softmax}\left(W_o h_i+b_o\right),
\end{equation}
where $W_o$ and $b_o$ represent the weights and biases of the output layer of $G(\Theta)$ , respectively, and $h_i$ is the hidden state of the $G(\Theta)$ at the $i^{th}$ time step, which incorporates information from both the image $I$ and the generated text tokens $t_{<i}$.

The learning objective in the finetuning generator stage is:
\begin{equation}
\mathcal{L}_{Cap}=-\sum_{i=1}^n \log P\left(t_i \mid I, t_{<i}\right).
\end{equation}
In the generation phase, we take a set of $N$ 3D medical images, $X=\{x_1, x_2, ..., x_N\}$. For each 3D volume $x_i$, we randomly sample a 2D slice as an input for the finetuned generator $G(\Theta)$, which then generates the textual description $T_i$ of 3D volume $x_i$. This can be mathematically formulated as
$
T_i=G(x_i|\Theta).
$
After the generation phase, we filter out duplicate descriptions and remove certain fixed vocabulary that lacks information using regular expressions. To enhance the accuracy and distinctiveness of the textual descriptions, we prepend the name of the dataset to the beginning of each description. Each 3D medical image is then paired with a synthesized textual description, forming the image-text pairs $D=\{(x_1,T_1),(x_2,T_2),...,(x_N,T_N)\}$ for subsequent representation learning. Examples of the generated text can be found in the appendix.

\subsection{3D Visual-Textual Representation Learning}
Given the dataset $\mathcal{D}$ of 3D medical images paired with synthetic text, we aim to learn the visual-textual representation via the image encoder be $f_I(\cdot)$ and the text encoder be $f_T(\cdot)$. The text encoder $f_T(\cdot)$, initialized with the weights from BioBERT \cite{lee2020biobert}, is frozen during pretraining to maximize computational efficiency during the extraction of text embeddings.

For a sample batch of image-text pairs $(X_i, T_i)$, we first compute their respective feature representations: $v_{e,i} = f_I(X_i)$ and $t_{e,i} = f_T(T_i)$.

we employ a contrastive learning objective to predict the matched pair $\left(v_{e, i}, t_{e, i}\right)$ among $N \times N$ potential image-text pairs, while concurrently ensuring that $N^2-N$ negative pairs are distinctly separated. Concretely, we utilize two non-linear visual and text projectors, $\mathcal{F}_l$ and $\mathcal{F}_v$, to transform $\mathbf{v}_{e, i}$ and $\mathbf{t}_{e, i}$ into the same dimensional space $d$, where $\hat{\mathbf{v}}_{e, i}=\mathcal{F}_I\left(\mathbf{v}_{e, i}\right)$, $\hat{\mathbf{t}}_{e, i}=\mathcal{F}_T\left(\mathbf{l}_{e, i}\right)$, and $\{\hat{\mathbf{v}}_{e, i}, \hat{\mathbf{t}}_{e, i}\} \in \mathbb{R}^d$. Subsequently, we generate image vectors $\left[\hat{\mathbf{V}}_{e, i}\right]_{i=1}^N$ and text vectors $\left[\hat{\mathbf{T}}_{e, i}\right]_{i=1}^N$ within a training batch to compute cosine similarities:

\begin{equation}
\mathcal{L}_v^{v 2 t}=-\log \frac{\exp \left(s_{i, i}^{v 2 t} / \sigma_1\right)}{\sum_{j=1}^K \exp \left(s_{i, j}^{v 2 t} / \sigma_1\right)}, ~\mathcal{L}_t^{t 2 v}=-\log \frac{\exp \left(s_{i, i}^{t 2 v} / \sigma_1\right)}{\sum_{j=1}^K \exp \left(s_{i, j}^{t 2 v} / \sigma_1\right)},
\end{equation}
where $\mathcal{L}_v^{v 2 t}$ and $\mathcal{L}_l^{t 2 v}$ are image-text and text-image InfoNCE \cite{oord2018representation} contrastive loss, respectively. $s_{i, i}^{v 2 t}=\hat{\mathbf{v}}_{e, i}^{\top} \hat{\mathbf{t}}_{e, i}$ and $s_{i, i}^{t 2 v}=\hat{\mathbf{t}}_{e, i}^{\top} \hat{\mathbf{v}}_{e, i}$ represent image-text and text-image similarities. $K$ is the batch size of each step. $\sigma_1$ is the temperature hyper-parameter set to 0.07 in our experiments.

The loss function can be articulated as:

\begin{equation}
\mathcal{L}_{{VLP}}=\frac{1}{2 N} \sum_{i=1}^N\left(\mathcal{L}_v^{v 2 t}+\mathcal{L}_t^{t 2 v}\right).
\end{equation}
Through overall loss $\mathcal{L}_{\mathrm{VLP}}$, the model learns maximal mutual information between the matched multi-modal pairs containing cross-view attributes within a batch.

\subsection{3D Visual Representation Learning}
\begin{figure}[t]
    \centering
    \includegraphics[width = 0.8\linewidth]{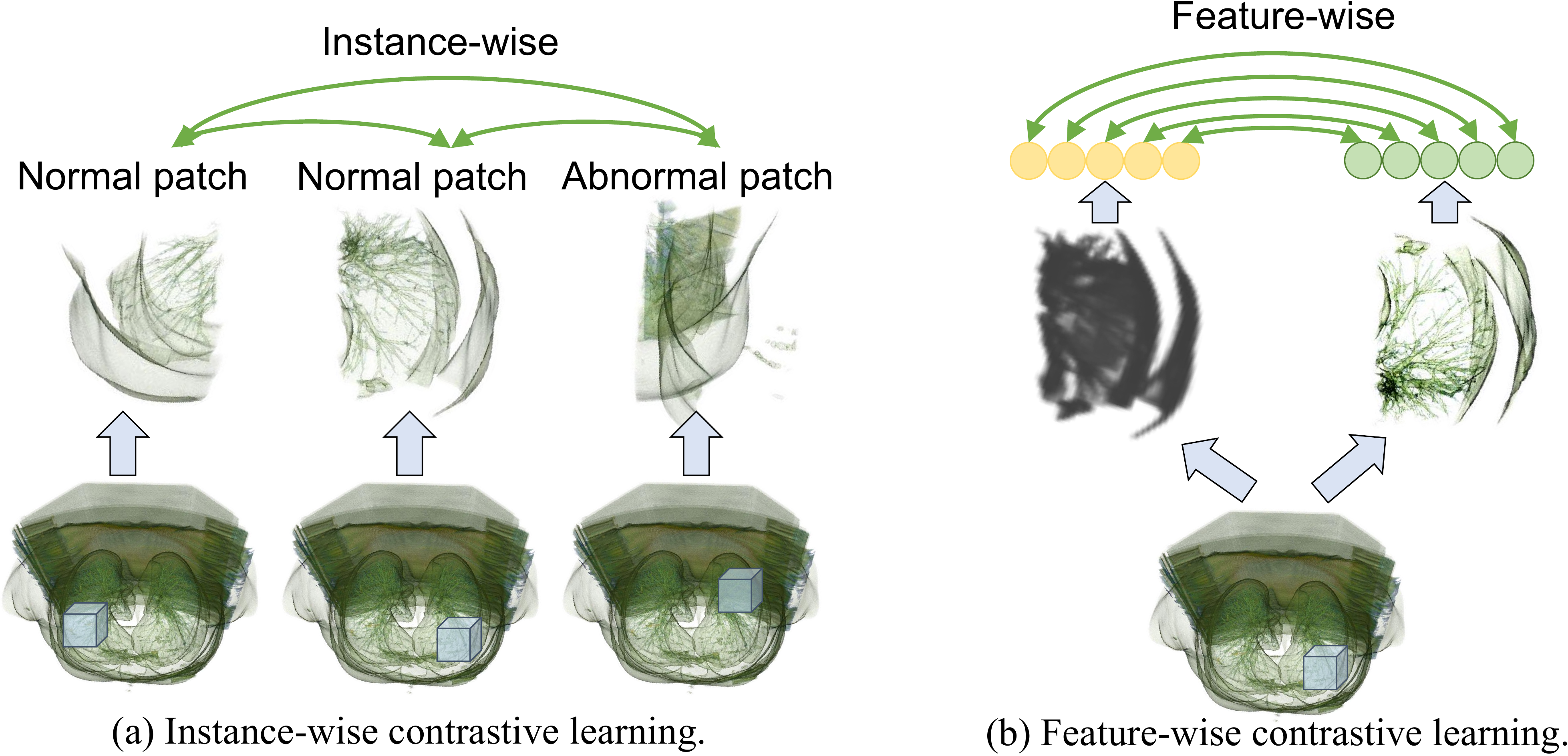}
    \caption{Graphically explanation of the bias introduced by instance-wise 3D image SSL and the effectiveness of our novel feature-wise 3D image SSL in mitigating bias arising from positive and negative sample pairs.}
    \label{fig:intance-wise}

\end{figure}

Image contrastive learning, commonly deployed to learn visual invariants, typically involves defining one positive sample (such as an augmented view) and treating the remainder of the batch's samples as negatives. Nevertheless, this rigid 1-to-n positive-negative pairing tends to introduce substantial bias when learning 3D visual representation, particularly because in 3D medical imaging, each sample represents a patch of the original volume. Consequently, as shown in Figure \ref{fig:intance-wise}, two slices could both represent normal organ semantics, even if their source volumes contain abnormal organs. Moreover, 1-to-n contrastive learning requires a large batch size~\cite{byol,barlowtwins}, which is not feasible in 3D visual learning tasks~\cite{liu2021swin}.

To address these challenges, we introduce a negative-free learning objective instead of the rigid positive-negative based loss. This objective aims to disentangle the latent space feature-wisely and maximize the information in each feature dimension \cite{barlowtwins}.

We first generate two distinct views $X^1$ and $X^2$ of the medical volume $X$ through random data augmentation. 
We initiate by normalizing the augmented embedding pairs $\{\mathbf{V}{\mathbf{e}}^{\mathbf{1}}, \mathbf{V}{\mathbf{e}}^{\mathbf{2}}\} \in \mathbb{R}^{N \times d}$ along the batch $K$ dimension. This normalization ensures each feature dimension has a zero-mean and $1 / \sqrt{K}$ standard deviation distribution, resulting in $\tilde{\mathbf{V}}_{\mathbf{e}}$. Subsequently, we compute their cross-correlation $\tilde{\mathbf{V}}_e^{\text {corr }}=\tilde{\mathbf{V}}_e^{\mathrm{1T}} \tilde{\mathbf{V}}_e^{\mathrm{2}}$. The following defines the feature-dimension decorrelation formulas:

where $N$ represents the batch size. Our objective is to minimize the off-diagonal elements of the cross-correlation matrix $\hat{V}^{corr}_e$ and maximize the diagonal elements. The loss function can be formulated as:

\begin{equation}
\small
\mathcal{L}_{VR}=  \frac{1}{D_{\prime}} \left\{
\underbrace{
\sum_{j}^{D^{\prime}}\left(1-\sum_i^K \mathbf{\tilde{V}}_{e, i}^{1,j \mathbf{T}} \mathbf{\tilde{V}}_{e, i}^{2, j}\right)^2
}_{\text{cross-view invariants}}
+ 
\underbrace{
\lambda_1 \sum_{j}^{D^{\prime}} \sum_{i \neq j}^{K} \mathbf{\tilde{V}}_{e, i}^{1,j \mathbf{T}} \mathbf{\tilde{V}}_{e, i}^{2, j}
}_{\text{cross-view superfluity  reduction}}
\right\}
, \quad
\tilde{\mathbf{V}_{e}}=\frac{\mathbf{V_{e}}-\mu_K(\mathbf{V}_{e})}{\sigma(\mathbf{V_{e}}) \sqrt{K}}.
\label{barlow loss}
\end{equation}

Here, $\lambda_1$ is a non-negative hyperparameter used to adjust the trade-off between learning invariants and reducing superfluity in Equation \ref{barlow loss}
We set the value of $\lambda_1$ according to the default setting used in~\cite{barlowtwins}.
The first term is crafted to learn a visual-invariant representation by optimizing the diagonal elements of the cross-correlation matrix $\hat{V}^{corr}_e$ to be close to one.
The second term is designed to lessen the correlation between distinct latent variables, thereby encouraging maximal information in each latent dimension by minimizing the off-diagonal elements in $\hat{V}^{corr}_e$.
Finally, the loss is normalized along the feature dimension $d$.

The overall loss function can be articulated as:
\begin{equation}
    \mathcal{L}=\lambda_1 \mathcal{L}_{VLP}+\lambda_2 \mathcal{L}_{VR},
\end{equation}
The coefficients $\lambda_1$ and $\lambda_2$ are used to control the weights, and we set $\lambda_1$ and $\lambda_2$ to be 1 and 0.01.
\section{Experiments}
\label{sec: experimients}
We conduct extensive experiments on a large number of cross-modal, cross-species, and cross-organ medical images. Generally speaking, image pretraining often encounters information bottlenecks, and the performance of downstream tasks does not improve with increasing amounts of data. However, due to the introduction of generated textual information, GTGM achieves good performance across a wide range of downstream tasks. In this section, we provide a detailed description of the pretraining and downstream task datasets, parameter settings, and report our experimental results.
\subsection{Dataset and Metrics}
\noindent\textbf{Dataset.\hspace{2mm}}Our dataset encompasses imaging data from three modalities: CT, MRI, and EM. The primary sources of CT and MRI data are the Medical Segmentation Decathlon (MSD) \cite{antonelli2022medical} competition, which includes 3D or 4D imaging of 10 different organs. During pretraining, we utilize all ImageTr and ImageTs, intentionally excluding labels. For the downstream segmentation tasks, we divide the ImageTr and corresponding ImageTs data into an 80\% training set and a 20\% test set. We conduct experiments using 1\%, 10\%, and 100\% of the training set data (excluding the 1\% setting when the data is insufficient to form a complete file). This allows us to evaluate the algorithm's performance under conditions of both label scarcity and abundance.

EM data primarily originate from large-scale EM datasets, namely the Full Adult Fly Brain (FAFB) \cite{schlegel2021automatic}, MitoEM \cite{wei2020mitoem}, FIB-25 \cite{takemura2017connectome}, and Kasthuri15 \cite{kasthuri2015saturated}. These datasets contain images from diverse organisms, including \textit{Drosophila}, mice, rats, and humans. For the downstream segmentation tasks, we evaluate the algorithm's performance using three datasets from the CREMI competition \cite{funke2016miccai}. The CREMI dataset consists of three subsets: A, B, and C, each containing 125 images. We select the last 50 images from each subset for testing, while training is conducted using either the first 75 images or the first 10 images from each subset.


\noindent\textbf{Metrics.\hspace{2mm}}Our primary evaluation of the algorithm's performance is conducted through downstream segmentation tasks. Among these, tasks involving CT and MRI scans fall into the category of semantic segmentation. Given the relatively small proportion of the entire volume occupied by the organs in these cases, we employ the Dice coefficient as a performance metric. In contrast, EM tasks are considered instance segmentation tasks. Here, there is no background in the volume and the neurons to be segmented are densely packed. Consequently, we use two metrics, Variation of Information (VOI) \cite{nunez2013machine} and Adjusted Rand Index (ARAND) \cite{arganda2015crowdsourcing}, to evaluate the segmentation performance.

\subsection{Implementation Details}
Our pretraining is conducted on 8 NVIDIA A100 GPUs, including both text-image and image-only pretraining. The batch size is set to 16 per GPU, with an initial learning rate of 2e-5 and a learning rate decay of 5e-2. All pretraining tasks are iterated for 100k iterations. For all downstream tasks, we use 2 NVIDIA RTX 3090 GPUs or 1 NVIDIA A100 GPU. For CT and MRI tasks, we train for 40k iterations, while for EM tasks we train for 100k iterations. We utilize the AdamW optimizer with beta coefficients set to 0.9 and 0.999 for all tasks.

\subsection{Experiment Results}
We conduct extensive experiments on downstream applications involving organ-wise, modality-wise, and species-wise segmentation tasks. During the pretraining phase, we train the vision encoder using the same dataset as described earlier. In the finetuning phase, we concurrently update the parameters of the pretrained vision encoder and a randomly initialized decoder, using various label ratios. Our framework is compared with state-of-the-art self-supervised algorithms, including BYOL \cite{byol}, BarlowTwins \cite{barlowtwins}, and SimSiam \cite{simsiam} for natural images, as well as the latest SOTA algorithm specifically designed for medical images, PCRLv2 \cite{pcrlv2}.
Due to the fact that these baselines either do not provide a 3D vision encoder~\cite{byol,barlowtwins,simsiam} or are only pretrained on limited 3D datasets~\cite{pcrlv2}, we replicate these algorithms on our pretraining dataset and evaluate their performance on downstream tasks using the same experimental setup.
To ensure compatibility with 3D medical images and enable a fair comparison, we adopt the 3D ResNet-50 \cite{he2016deep} as the vision encoder in both the pretraining and fine-tuning stages for all experiments. For pixel-level instance segmentation in downstream tasks, we employ a U-Net-style decoder.
The results obtained using the Swin-Transformer-base vision encoder \cite{liu2021swin} can be found in the appendix.

\paragraph{Experimental Results for CT Segmentation.}
CT imaging plays a crucial role in the medical field, particularly in lesion segmentation. However, the limited contrast differences between different tissues in CT images can lead to blurry boundaries between lesions and surrounding normal tissues. The segmentation results of six datasets of CT images are presented in Table \ref{tab:CT1}. We achieve state-of-the-art (SOTA) performance on all datasets except for Hepatic Vessel. This discrepancy may be attributed to the fact that in the pretraining dataset, the Hepatic Vessel often coexists with the liver, and textual descriptions tend to focus more on the larger scale of the liver itself, resulting in a decline in the effectiveness of pretraining.
\begin{table}[t]
\centering
\caption{Mean Dice scores (\%) of CT image segmentation results. {\color[HTML]{FF0000} Red} and {\color[HTML]{007ACC} blue} entries denote the best and second-best results, respectively.}
\label{tab:CT1}
\begin{tabular}{lcccccccc}
\toprule[1.2pt]
                         & \multicolumn{3}{c}{Liver}                                                          & \multicolumn{3}{c}{Pancreas}                                                       & \multicolumn{2}{c}{Lung}                            \\
\multirow{-2}{*}{Method} & 1 \%                         & 10\%                         & 100\%                         & 1\%                          & 10\%                         & 100\%                         & 10\%                         & 100\%                         \\ \hline \hline
Random                   & 45.21                        & 51.24                        & 61.34                        & 37.07                        & 56.21                        & 63.96                        & 57.36                        & 73.49                        \\ \hline
BYOL \cite{byol}                    & 45.11                        & 52.33                        & 61.67                        & 39.83                        & 56.8                         & 64.51                        & {\color[HTML]{00B0F0} 59.84} & 76.41                        \\
SimSiam \cite{simsiam}                 & 48.22                        & 51.29                        & 62.39                        & {\color[HTML]{00B0F0} 40.03} & 54.82                        & {\color[HTML]{00B0F0} 64.69} & 59.71                        & {\color[HTML]{00B0F0} 79.43} \\
BarlowTwins \cite{barlowtwins}             & 50.13                        & 55.85                        & 64.93                        & 39.67                        & {\color[HTML]{00B0F0} 57.01} & 63.59                        & 55.22                        & 71.37                        \\
PCRLv2 \cite{pcrlv2}                 & {\color[HTML]{00B0F0} 51.69} & {\color[HTML]{00B0F0} 56.63} & {\color[HTML]{00B0F0} 65.19} & 39.80                        & 56.05                        & 63.38                        & 55.30                        & 74.19                        \\ \hline \hline
GTGM                     & {\color[HTML]{FF0000} 52.46} & {\color[HTML]{FF0000} 58.67} & {\color[HTML]{FF0000} 65.61} & {\color[HTML]{FF0000} 40.55} & {\color[HTML]{FF0000} 59.96} & {\color[HTML]{FF0000} 65.61} & {\color[HTML]{FF0000} 61.3}  & {\color[HTML]{FF0000} 80.19} \\ \bottomrule[1.2pt]
\end{tabular}

\end{table}

\begin{table}[t]
\centering
\begin{tabular}{lcccccccc}
\toprule[1.2pt]
                         & \multicolumn{3}{c}{HepaticVessel}                                                   & \multicolumn{3}{c}{Colon}                                                           & \multicolumn{2}{c}{Spleen}                          \\
\multirow{-2}{*}{Method} & 1 \%                         & 10\%                         & 100\%                         & 1\%                          & 10\%                         & 100\%                         & 10\%                         & 100\%                         \\ \hline \hline
Random                   & 49.84                        & 51.53                        & 64.56                        & 50.29                        & 50.8                         & 50.6                         & 73.85                        & 84.92                        \\ \hline
BYOL \cite{byol}                     & 49.67                        & 58.85                        & {\color[HTML]{FF0000} 65.57} & 50.1                         & 50.29                        & 50.22                        & 77.64                        & 85.98                        \\
SimSiam \cite{simsiam}                  & 50.07                        & 52.34                        & 63.78                        & 50.27                        & 51.18                        & {\color[HTML]{00B0F0} 53.8}  & 81.93                        & 83.49                        \\
BarlowTwins \cite{barlowtwins}              & {\color[HTML]{FF0000} 51.08} & {\color[HTML]{00B0F0} 59.21} & 64.77                        & {\color[HTML]{00B0F0} 50.62} & {\color[HTML]{00B0F0} 51.46} & 51.61                        & {\color[HTML]{00B0F0} 86.43} & {\color[HTML]{00B0F0} 87.91} \\
PCRLv2 \cite{pcrlv2}                 & 50.12                        & 58.82                        & 64.97                        & 50.08                        & 51.43                        & 53.13                        & 84.32                        & 85.12                        \\ \hline \hline
GTGM                     & {\color[HTML]{00B0F0} 50.39} & {\color[HTML]{FF0000} 59.74} & {\color[HTML]{00B0F0} 65.13} & {\color[HTML]{FF0000} 51.12} & {\color[HTML]{FF0000} 51.74} & {\color[HTML]{FF0000} 53.88} & {\color[HTML]{FF0000} 86.95} & {\color[HTML]{FF0000} 89.64} \\ \bottomrule[1.2pt]
\end{tabular}

\end{table}

\paragraph{Experimental Results for MRI Segmentation.}
In comparison to CT imaging, MRI imaging typically involves four dimensions and exhibits lower imaging resolution, as well as more artifacts and noise in the images. Consequently, lesion segmentation in MRI images presents greater challenges. In our pretraining approach, tailored to 3D imaging, we extract the last three dimensions from the MRI images as input to the network. Despite these challenges, our approach has delivered promising experimental results, as depicted in Table \ref{tab:MRI1}. Consistently, our approach achieves either optimal or near-optimal solutions on the MRI dataset. We observe that numerous image-based self-supervised approaches yield degraded outcomes (with Dice scores lower than random initialization) due to variations in image dimensions within the MRI dataset. However, our approach exhibits robustness and effectively mitigates the degradation of the pretrained network by incorporating text as guidance.

\begin{table}[t]
\centering
\caption{Mean Dice scores (\%) of MRI image segmentation results. {\color[HTML]{FF0000} Red} and {\color[HTML]{007ACC} blue} entries denote the best and second-best results, respectively.}
\label{tab:MRI1}
\scalebox{0.9}{
\begin{tabular}{lcccccccccc}
\toprule[1.2pt]
                         & \multicolumn{3}{c}{BrainTumour}                                                      & \multicolumn{2}{c}{Heart}                            & \multicolumn{3}{c}{Hippocampus}                                                     & \multicolumn{2}{c}{Prostate}                         \\
\multirow{-2}{*}{Method} & 1\%                          & 10\%                         & 100\%                         & 10\%                         & 100\%                         & 1\%                          & 10\%                         & 100\%                         & 10\%                         & 100\%                         \\ \hline \hline
Random                   & 30.33                        & 32.37                        & 40.45                        & 82.37                        & 94.82                        & 48.32                        & 78.46                        & 84.18                        & 33.53                        & 39.19                        \\ \hline
BYOL \cite{byol}                     & 31.89                        & 31.95                        & 44.94                        & 83.31                        & {\color[HTML]{FF0000} 94.97} & 50.96                        & {\color[HTML]{FF0000} 79.82} & {\color[HTML]{007ACC} 84.47} & 40.73                        & {\color[HTML]{007ACC} 43.72} \\
SimSiam \cite{simsiam}                  & 31.27                        & 32.67                        & 37.45                        & {\color[HTML]{007ACC} 86.06} & 93.51                        & 52.24                        & 78.99                        & 83.35                        & {\color[HTML]{FF0000} 42.48} & 41.64                        \\
BarlowTwins \cite{barlowtwins}              & 32.13                        & 32.21                        & {\color[HTML]{FF0000} 45.83} & 83.37                        & 94.68                        & 50.84                        & 78.75                        & 83.96                        & 33.37                        & 40.15                        \\
PCRLv2 \cite{pcrlv2}                & {\color[HTML]{007ACC} 32.83} & {\color[HTML]{FF0000} 34.87} & 43.14                        & 85.89                        & 90.77                        & {\color[HTML]{007ACC} 52.29} & 77.38                        & 81.24                        & 32.73                        & 40.54                        \\ \hline \hline
GTGM                     & {\color[HTML]{FF0000} 33.19} & {\color[HTML]{007ACC} 34.12} & {\color[HTML]{007ACC} 45.23} & {\color[HTML]{FF0000} 86.33} & {\color[HTML]{007ACC} 94.71} & {\color[HTML]{FF0000} 53.41} & {\color[HTML]{007ACC} 79.01} & {\color[HTML]{FF0000} 84.92} & {\color[HTML]{007ACC} 40.93} & {\color[HTML]{FF0000} 44.24} \\ \bottomrule[1.2pt]
\end{tabular}}

\end{table}


\paragraph{Experimental Results for EM Neuron Segmentation.}
Electron Microscopy (EM) is an imaging technique with a resolution approximately a thousand times greater than CT and MRI, permitting the examination of structures at the nano and sub-nanometer levels. The ultra-high resolution makes neuron segmentation tasks in EM particularly challenging due to the densely packed structures.
The typical methodology for EM neuron segmentation involves neural network-based affinity prediction, followed by post-processing with methods like WaterZ \cite{funke2018large} for instance segmentation. Due to the complexity of neurons, commonly used Vision-Language Pretraining (VLP) methods are not applicable. However, our proposed GTGM overcomes this limitation and demonstrates the effectiveness of generated text as a form of self-supervision training guidance. GTGM achieves state-of-the-art results on three neuron datasets in two settings. Please refer to Table \ref{tab:emSeg} for specific numerical results. 

\begin{figure}[t]
    \centering
    \includegraphics[width = \linewidth]{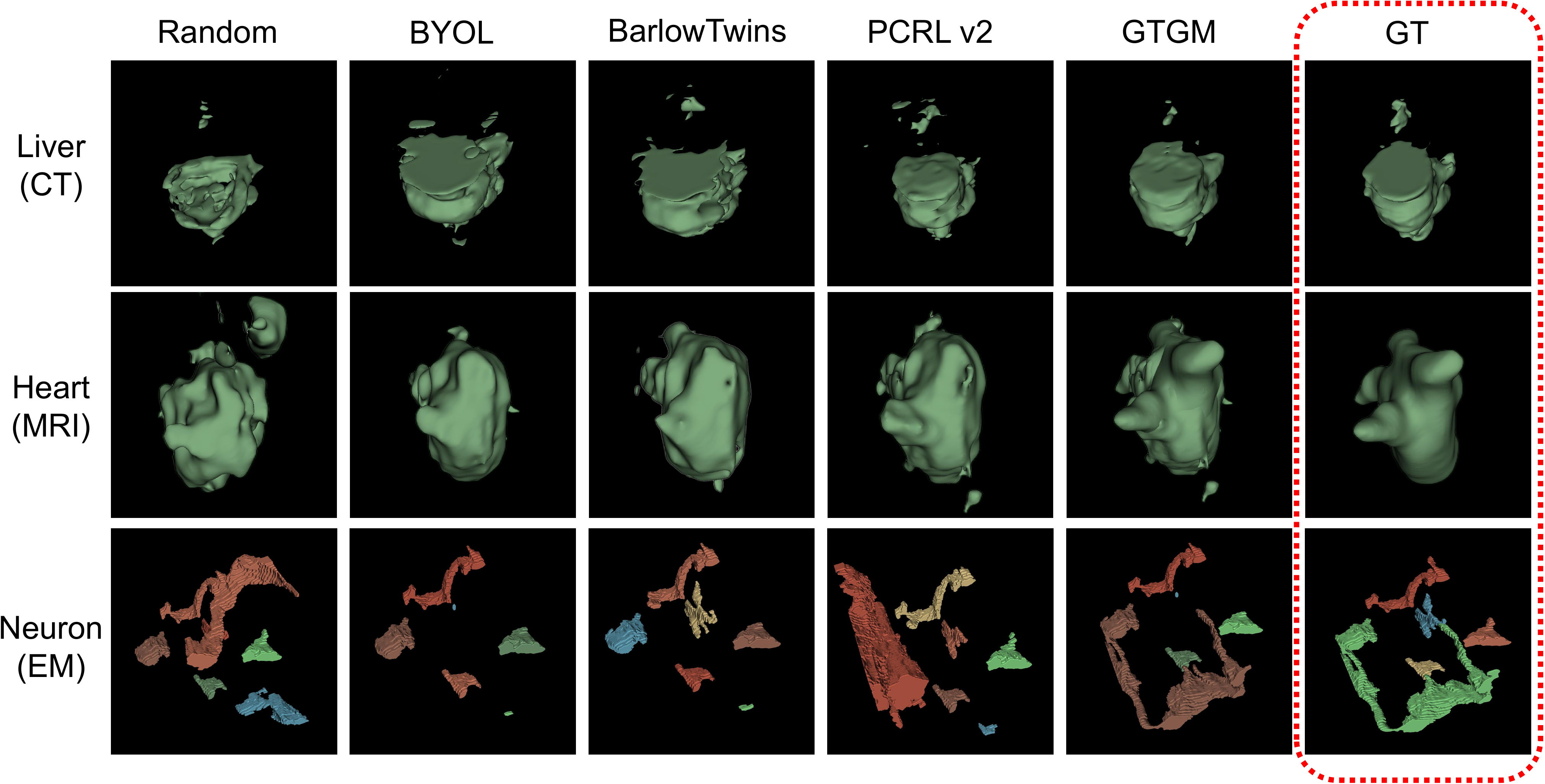}
    \caption{Visualization Results of 3D Instance Segmentation. `GT' indicates the ground truth (Neurons are distinguished by their geometric shapes rather than color labels for the same instance).}
    \label{fig:visual}
\end{figure}

\paragraph{Visual Results.}
GTGM demonstrates significant improvements in medical instance segmentation, particularly in terms of the connectivity of segmented surfaces, as shown in Figure \ref{fig:visual}. Our approach effectively segments liver tumors, and surface geometric structures of the left atrium, and exhibits the strongest integrity in neuronal segmentation.

\begin{table}[t]
\centering
\caption{Neuron segmentation results of three EM datasets. {\color[HTML]{FF0000} Red} and {\color[HTML]{007ACC} blue} entries denote the best and second-best results, respectively (The performance is better when the values of VOI and Arand are smaller).}
\scalebox{0.75}{
\label{tab:emSeg}
\begin{tabular}{lcccccccccccc}
\toprule[1.2pt]
                         & \multicolumn{2}{c}{CREMI A 10}                              & \multicolumn{2}{c}{CREMI A 75}                              & \multicolumn{2}{c}{CREMI B 10}                              & \multicolumn{2}{c}{CREMI B 75}                              & \multicolumn{2}{c}{CREMI C 10}                              & \multicolumn{2}{c}{CREMI C 75}                              \\
\multirow{-2}{*}{Method} & VOI                          & Arand                        & VOI                          & Arand                        & VOI                          & Arand                        & VOI                          & Arand                        & VOI                          & Arand                        & VOI                          & Arand                        \\ \hline \hline
Random                   & 1.051                        & 0.184                        & 0.744                        & 0.104                        & 2.181                        & 0.234                        & 1.560                        & 0.261                        & 1.987                        & {\color[HTML]{007ACC} 0.145} & 1.424                        & 0.140                        \\ \hline
BYOL \cite{byol}                     & 0.961                        & 0.206                        & 0.764                        & 0.119                        & 1.581                        & 0.155                        & 1.441                        & {\color[HTML]{007ACC} 0.142} & 1.672                        & 0.196                        & 1.326                        & {\color[HTML]{007ACC} 0.124} \\
SimSiam \cite{simsiam}                  & 0.985                        & {\color[HTML]{007ACC} 0.171} & 0.770                        & 0.100                        & {\color[HTML]{007ACC} 1.511} & {\color[HTML]{007ACC} 0.125} & 1.332                        & 0.150                        & 1.578                        & 0.178                        & 1.364                        & 0.130                        \\
BarlowTwins \cite{barlowtwins}              & 0.987                        & 0.200                        & 0.743                        & 0.101                        & 1.584                        & 0.185                        & {\color[HTML]{007ACC} 1.291} & 0.187                        & {\color[HTML]{007ACC} 1.483} & 0.147                        & {\color[HTML]{007ACC} 1.303} & 0.129                        \\
PCRLv2 \cite{pcrlv2}                 & {\color[HTML]{007ACC} 0.921} & 0.189                        & {\color[HTML]{007ACC} 0.738} & {\color[HTML]{007ACC} 0.100} & 1.568                        & 0.158                        & 1.374                        & 0.157                        & 1.596                        & 0.178                        & 1.326                        & 0.127                        \\ \hline \hline
GTGM                    & {\color[HTML]{FF0000} 0.902} & {\color[HTML]{FF0000} 0.166} & {\color[HTML]{FF0000} 0.728} & {\color[HTML]{FF0000} 0.092} & {\color[HTML]{FF0000} 1.525} & {\color[HTML]{FF0000} 0.117} & {\color[HTML]{FF0000} 1.279} & {\color[HTML]{FF0000} 0.106} & {\color[HTML]{FF0000} 1.422} & {\color[HTML]{FF0000} 0.137} & {\color[HTML]{FF0000} 1.280} & {\color[HTML]{FF0000} 0.118} \\ \bottomrule[1.2pt]
\end{tabular}}

\end{table}

\section{Further Analysis}
\subsection{Ablation Study of Component Design}
Table \ref{aba:framewark} presents the results of the ablation study for our proposed three components. We utilize 3D ResNet-50~\cite{he2016deep} as the vision backbone and conduct instance segmentation tests on three representative datasets (Liver, Prostate, CREMI C).
As shown in Table \ref{aba:framewark}, the impact of utilizing the synthetic text-guided VLP is clearly evident in the significant improvements observed in downstream tasks. The combination of all components of GTGM achieves the highest performance, clearly demonstrating the efficacy of GTGM. Learning only visual invariants or visual-textual invariants exhibits varying performance across different datasets.
Notably, only learning visual invariants during pretraining proves to be more effective for large-scale and dense instance segmentation, while visual-textual invariants excel in guiding small and sparse segmentation. This difference in performance can be attributed to the fact that the former captures more intricate structural information, which is challenging to concisely describe through text alone.
\begin{table}[hb]
\centering
\caption{Ablation study of our framework is conducted on CT, MRI, and EM datasets, reporting Dice scores for CT and MRI, and VOI results for EM. {\color[HTML]{FF0000}Red} and {\color[HTML]{007ACC}blue} entries indicate the best and second-best results, respectively.}
\label{aba:framewark}
\scalebox{0.9}{
\begin{tabular}{cccccccccc}
\toprule[1.2pt]
\multicolumn{3}{c}{Training tasks}           & \multicolumn{3}{c}{Liver (Dice $\uparrow$)}                                                              & \multicolumn{2}{c}{Prostate (Dice $\uparrow$)}                                                & \multicolumn{2}{c}{CREMI C (VOI $\downarrow$)}                           \\
$\mathcal{L}_{Cap}$ & $\mathcal{L}_{VLP}$ & \multicolumn{1}{c|}{$\mathcal{L}_{VR}$} & 1 \%                          & 10\%                          & 100\%                          & 10\%                          & \multicolumn{1}{c|}{100\%}                          & 10                           & 75                           \\ \midrule \midrule
       &        & \multicolumn{1}{c|}{$\checkmark$}    & 50.11                         & 55.91                         & {\color[HTML]{007ACC} 64.87} & 33.37                         & \multicolumn{1}{c|}{40.15}                         & {\color[HTML]{007ACC} 1.483} & {\color[HTML]{007ACC} 1.303} \\
       & $\checkmark$     & \multicolumn{1}{c|}{}      & 46.84                         & 51.95                         & 61.21                         & 33.67                         & \multicolumn{1}{c|}{39.21}                         & 1.871                        & 1.413                        \\
$\checkmark$     & $\checkmark$     & \multicolumn{1}{c|}{}      & {\color[HTML]{007ACC} 51.89} & {\color[HTML]{007ACC} 57.63} & 64.39                         & {\color[HTML]{007ACC} 38.91} & \multicolumn{1}{c|}{{\color[HTML]{007ACC} 41.39}} & 1.497                        & 1.333                        \\
$\checkmark$     & $\checkmark$     & \multicolumn{1}{c|}{$\checkmark$}    & {\color[HTML]{FF0000} 52.46}  & {\color[HTML]{FF0000} 58.67}  & {\color[HTML]{FF0000} 65.61}  & {\color[HTML]{FF0000} 40.93}  & \multicolumn{1}{c|}{{\color[HTML]{FF0000} 44.24}}  & {\color[HTML]{FF0000} 1.422} & {\color[HTML]{FF0000} 1.280} \\ \bottomrule[1.2pt]
    \end{tabular}}

\end{table}
\begin{table}[t]
\begin{minipage}[t]{0.48\textwidth}
\centering
\caption{Analysis of the trade-off between pretraining and finetuning.}
\label{tab:trade}
\scalebox{0.9}{
\begin{tabular}{c|ccc}
\toprule[1.2pt]
\multirow{2}{*}{Iters} & \multicolumn{3}{c}{Pancreas}                     \\
                       & 1 \%           & 10\%           & 100\%           \\ \midrule  \midrule
3.5 k                  & 38.93          & 56.7           & 64.73          \\
7 k                    & 39.26          & 56.09          & 64.97          \\
15 k                   & 39.67          & 56.83          & 64.95          \\
50 k                   & 40.13          & \textbf{56.96} & \textbf{65.61} \\
Last                   & \textbf{40.55} & 56.87          & 65.47          \\ \bottomrule[1.2pt]
\end{tabular}}
\end{minipage}
\hfill
\begin{minipage}[t]{0.48\textwidth}

\centering
\caption{Error bars of our methods across three modalities.}
\label{tab:error}
\scalebox{0.9}{
\begin{tabular}{ccc}
\toprule[1.2pt]
Dataset       & 10 \%            & 100\%            \\ \midrule \midrule
Heart  & 86.33 $\pm$ 0.41  & 94.71 $\pm$ 0.39  \\
Spleen & 86.95 $\pm$ 0.29  & 89.64 $\pm$ 0.43  \\ \bottomrule[1.2pt]
       &                  &                  \\ \toprule[1.2pt]
Metrics       & CREMI C 10       & CREMI C 75       \\ \midrule \midrule
VOI    & 1.422 $\pm$ 0.031 & 1.28 $\pm$ 0.025  \\
Arand  & 0.137 $\pm$ 0.011 & 0.118 $\pm$ 0.008 \\ \bottomrule[1.2pt]
\end{tabular}}
\end{minipage}

\end{table}

\subsection{Analysis of the Trade-off between Pretrain and Downstream tasks}
The difference in objective functions between the pretraining and finetuning phases can lead to suboptimal performance in downstream tasks, despite the convergence of the loss function during pretraining. 
This highlights the existence of a trade-off between these two stages. To showcase this trade-off, we conduct experiments on the representative Pancreas dataset, and the results are presented in Table \ref{tab:trade}.
Notably, the bold values in the table indicate the optimal segmentation results obtained under the current settings. These results signify that the number of iterations in our pretraining phase closely aligns with the performance achieved in the downstream task. This observation highlights the strong alignment between our objective function design and the requirements of the downstream task.

\subsection{Analysis of Error Bars}

Table \ref{tab:error} presents the error bars of our segmentation results on three modalities. We conduct three runs for each task and compute the mean and standard deviation of results. As observed from Table \ref{tab:error}, our results demonstrate relatively minor variations, indicating the stability of GTGM's performance in downstream tasks.

\section{Conclusion}

This work presents GTGM, a generative text-guided 3D vision-language pretraining framework. It accomplishes both instance-level visual-textual alignment and feature-level visual representation alignment using only 3D medical image inputs. GTGM delivers outstanding performance on 13 diverse medical datasets, tackling a variety of segmentation tasks with different data ratios. This demonstrates the efficiency and effectiveness of GTGM. 
Our work not only achieves the best performance but also opens up new opportunities to apply VLP to 3D medical images without relying on paired text.
The broader impact and limitations are shown in the Appendix.



\newpage
\appendix


\section{Overview}
In the supplementary material, we provide detailed explanations of our models and discuss the technical aspects involved. We have also included additional experimental results showcasing the performance of the Transformer backbone. Moreover, we have included numerous visualizations to enhance the understanding of our approach. To facilitate readability, we have also provided pseudo code for the core procedures.

\section{Latent Representation of Pretrained Model}
\begin{figure}[htb]
    \centering
    \includegraphics[width = \linewidth]{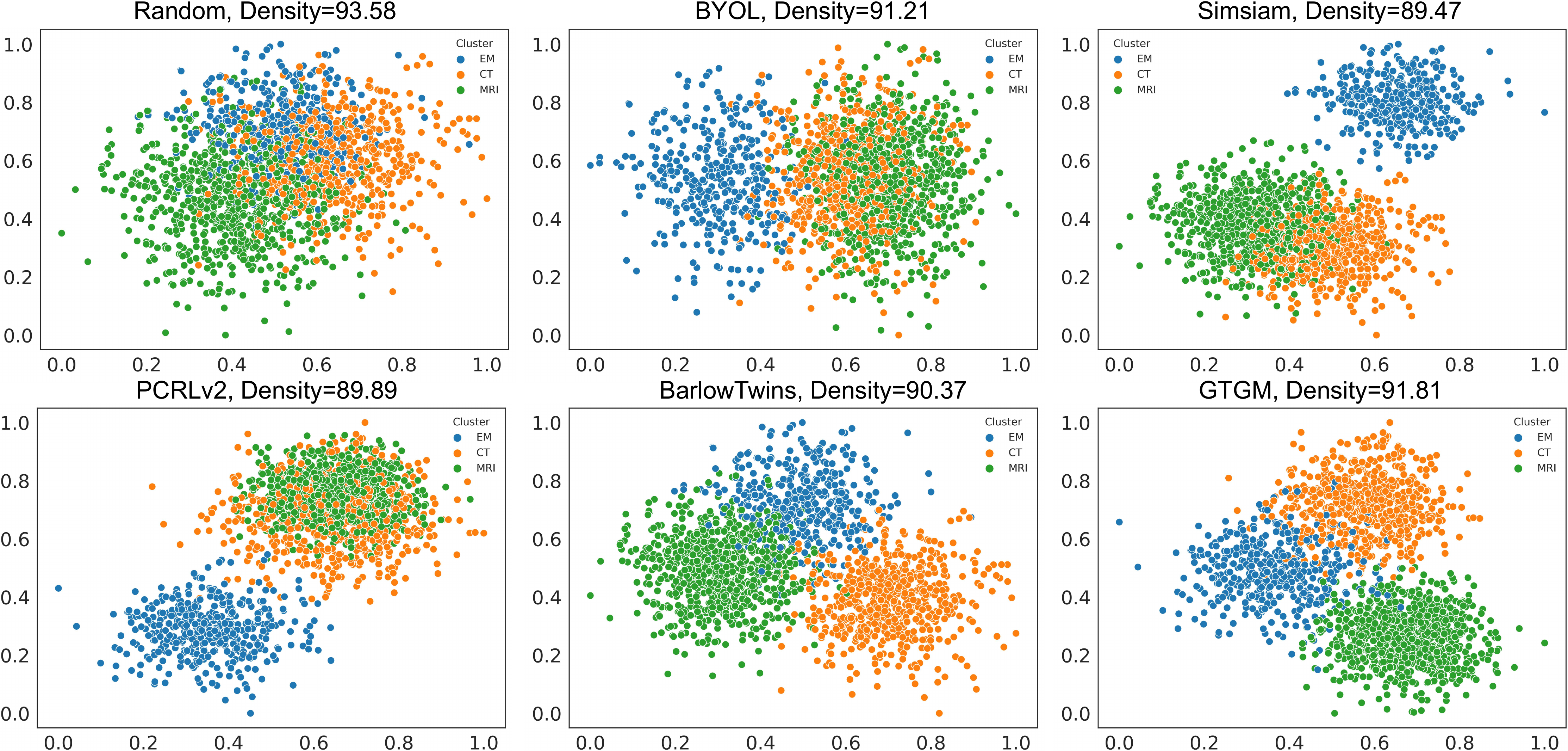}
    \caption{2d Visualization of volume Representation. All visualizations are rendered using t-SNE.}
    \label{fig:tsne}
\end{figure}

We implement various pretraining approaches~\cite{pcrlv2,barlowtwins,byol,simsiam} and GTGM on diverse 3D medical image datasets with annotation. Then we utilize the pretrained visual encoder to extract the latent representation from three distinct medical image modalities, namely CT, MRI, and EM. Subsequently, we subject the representation to dimensionality reduction using the t-SNE algorithm~\cite{van2008visualizing}. The resulting 2D representations obtained from different models are depicted in Figure \ref{fig:tsne}.

Based on the visualization results obtained from t-SNE in Figure \ref{fig:tsne}, it is evident that in scenarios with multiple modalities, the representation extracted through pretraining tend to exhibit confusion. Specifically, the CT and MRI modalities, compared to EM, display significantly lower resolution by several orders of magnitude. Although most pretraining models can effectively discriminate EM from the other two modalities, distinguishing between CT and MRI becomes challenging. This observation provides an explanation for the occurrence of model collapse phenomena in our experiments, wherein pretrained weights sometimes perform worse than random initialization.

Among all the models evaluated, only BarlowTwins and GTGM demonstrate effective differentiation among the three modalities. To further investigate this, we compute the density of the 2D representation, represented as the reciprocal of the k-nearest neighbor distances (higher values indicating tighter clustering). Interestingly, our model exhibits superior class separability compared to BarlowTwins, indicating that our text-guided approach captures informative features more effectively. In contrast, BarlowTwins' representation tend to be sparser, leading to potential overlap among modalities. This finding demonstrates the efficacy of our generative text-guided framework in capturing discriminated aspects among the three data modalities.

The additional analysis on density further supports the superiority of our approach in terms of class separability. The density measurements confirm that our model achieves tighter clustering, which is beneficial for class discrimination and avoiding confounding among the modalities. Overall, these findings highlight the effectiveness of our approach in capturing informative features and facilitating discrimination among the three different data types, surpassing the performance of the BarlowTwins method.
\section{Discussion}
\subsection{Limitation}
In the context of ablation experiments discussed in the main text, it was observed that for large-scale or higher-dimensional datasets, utilizing text generated through slicing for text-to-image pretraining yielded marginal information gain. Conversely, pretraining based on multi-view features derived from images proved more effective. In the future, it may be worthwhile to consider augmenting textual data appropriately or generating additional alternative texts through multiple slicing techniques, thereby harnessing the information contained within the text more comprehensively.

\subsection{Broader Impact}
The potential of vision-language multimodal pretraining has been widely recognized. Our proposed generation approach offers the possibility of joint training for datasets lacking textual descriptions. In addition to medical datasets, this approach can be extended to train on challenging datasets such as videos, point clouds, and light fields that are difficult to describe. Furthermore, directly fusing multimodal features at the downstream stage has been shown to significantly improve model performance. For instance, \cite{liu2023clip} achieved remarkable performance gains by introducing simple text prompts at the downstream stage, although the use of generated text to directly guide downstream tasks has yet to be validated. 

Moreover, our approach provides a means to leverage large language models (LLMs) effectively in computer vision tasks, leveraging existing pretrained weights. In the future, exploring the synchronization of LLM models with techniques like Stable Diffusion can be pursued to achieve zero-cost acquisition of high-quality data through text-guided image generation.

\subsection{Future Work}
Investigating these methodologies on varied medical data types, such as electrocardiograms linked with clinical monitoring records and multilingual reports, presents a fascinating future trajectory~\cite{li2023frozen,wan2023med}. Moreover, the alignment of heterogeneous modality data can be perceived as a data fusion task, an issue frequently tackled in the field of physics~\cite{cheng2023machine,cheng2022data,liu2022enkf} or recommendation system~\cite{Wan2022SpatioTemporalCL} .

\section{Transformer-based Results}
To validate the robustness and stability of our approach, we conducted experiments using a Transformer-based backbone. Taking inspiration from the network architecture of SwinUNETR \cite{tang2022self}, we fine-tuned our model on an electron microscopy dataset. The experimental results, as shown in Table \ref{tab:EM}, demonstrate the effectiveness of our approach.

Based on our experimental results, it can be observed that the performance of using Swin Transformer as a backbone is slightly inferior to that of using ResNet50 as a backbone. Additionally, the segmentation results of Swin Transformer are worse when dealing with a small amount of data. However, in comparison to the ResNet backbone, the gains obtained from pretraining for downstream tasks are more significant. Therefore, when larger datasets are available, Swin Transformer as a backbone holds tremendous potential.

\begin{table}[htb]
\centering
\caption{Experimental Results of Swin Transformer as a Backbone for Electron Microscope Neuron Segmentation. {\color[HTML]{FF0000} Red} and {\color[HTML]{007ACC} blue} entries denote the best and second-best results, respectively. Among these models, SwinUNETR refers to the implementation of the original pretraining approach described in the respective paper, with results reproduced on our pretraining dataset.}
\label{tab:EM}
\scalebox{0.75}{
\begin{tabular}{lcccccccccccc}
\toprule[1.2pt]
                         & \multicolumn{2}{c}{CREMI A 10}                              & \multicolumn{2}{c}{CREMI A 75}                              & \multicolumn{2}{c}{CREMI B 10}                              & \multicolumn{2}{c}{CREMI B 75}                              & \multicolumn{2}{c}{CREMI C 10}                              & \multicolumn{2}{c}{CREMI C 75}                              \\
\multirow{-2}{*}{Method} & VOI                          & Arand                        & VOI                          & Arand                        & VOI                          & Arand                        & VOI                          & Arand                        & VOI                          & Arand                        & VOI                          & Arand                        \\ \hline \hline
Random                   & 3.720                        & 0.898                        & 0.967                        & 0.244                        & 4.495                        & 0.646                        & 1.988                        & 0.289                        & 5.082                        & 0.770                        & 1.537                        & 0.175                        \\ \hline
BYOL \cite{byol}                     & 1.943                        & 0.559                        & 0.894                        & 0.220                        & 4.400                        & 0.670                        & 1.902                        & 0.256                        & 2.570                        & 0.448                        & 1.539                        & 0.210                        \\
SimSiam \cite{simsiam}                  & {\color[HTML]{00B0F0} 1.832} & {\color[HTML]{00B0F0} 0.531} & 0.927                        & 0.192                        & 3.381                        & 0.442                        & 1.871                        & 0.282                        & 2.419                        & 0.428                        & 1.442                        & 0.166                        \\
BarlowTwins \cite{barlowtwins}              & 2.104                        & 0.615                        & 0.956                        & 0.194                        & 3.292                        & 0.590                        & 1.851                        & {\color[HTML]{00B0F0} 0.201} & 2.419                        & 0.434                        & 1.437                        & 0.162                        \\
PCRLv2 \cite{pcrlv2}                 & 2.094                        & 0.640                        & 0.881                        & 0.184                        & {\color[HTML]{00B0F0} 3.174} & {\color[HTML]{00B0F0} 0.442} & {\color[HTML]{00B0F0} 1.594} & 0.241                        & {\color[HTML]{00B0F0} 2.219} & {\color[HTML]{FF0000} 0.298} & 1.474                        & 0.164                        \\
SwinUNETR \cite{tang2022self}                & 1.913                        & 0.579                        & {\color[HTML]{00B0F0} 0.855} & {\color[HTML]{FF0000} 0.177} & 3.813                        & 0.617                        & 1.859                        & 0.208                        & 2.419                        & 0.434                        & {\color[HTML]{00B0F0} 1.423} & {\color[HTML]{00B0F0} 0.160} \\ \hline \hline
GTGM                   & {\color[HTML]{FF0000} 1.693} & {\color[HTML]{FF0000} 0.529} & {\color[HTML]{FF0000} 0.832} & {\color[HTML]{00B0F0} 0.180} & {\color[HTML]{FF0000} 2.949} & {\color[HTML]{FF0000} 0.419} & {\color[HTML]{FF0000} 1.423} & {\color[HTML]{FF0000} 0.160} & {\color[HTML]{FF0000} 2.188} & {\color[HTML]{00B0F0} 0.304} & {\color[HTML]{FF0000} 1.393} & {\color[HTML]{FF0000} 0.151} \\ \bottomrule[1.2pt]
\end{tabular}}

\end{table}

\section{Visualization}
\subsection{Visualization of Segmentation Results}
\begin{figure}[htb]
    \centering
    \includegraphics[width=\linewidth]{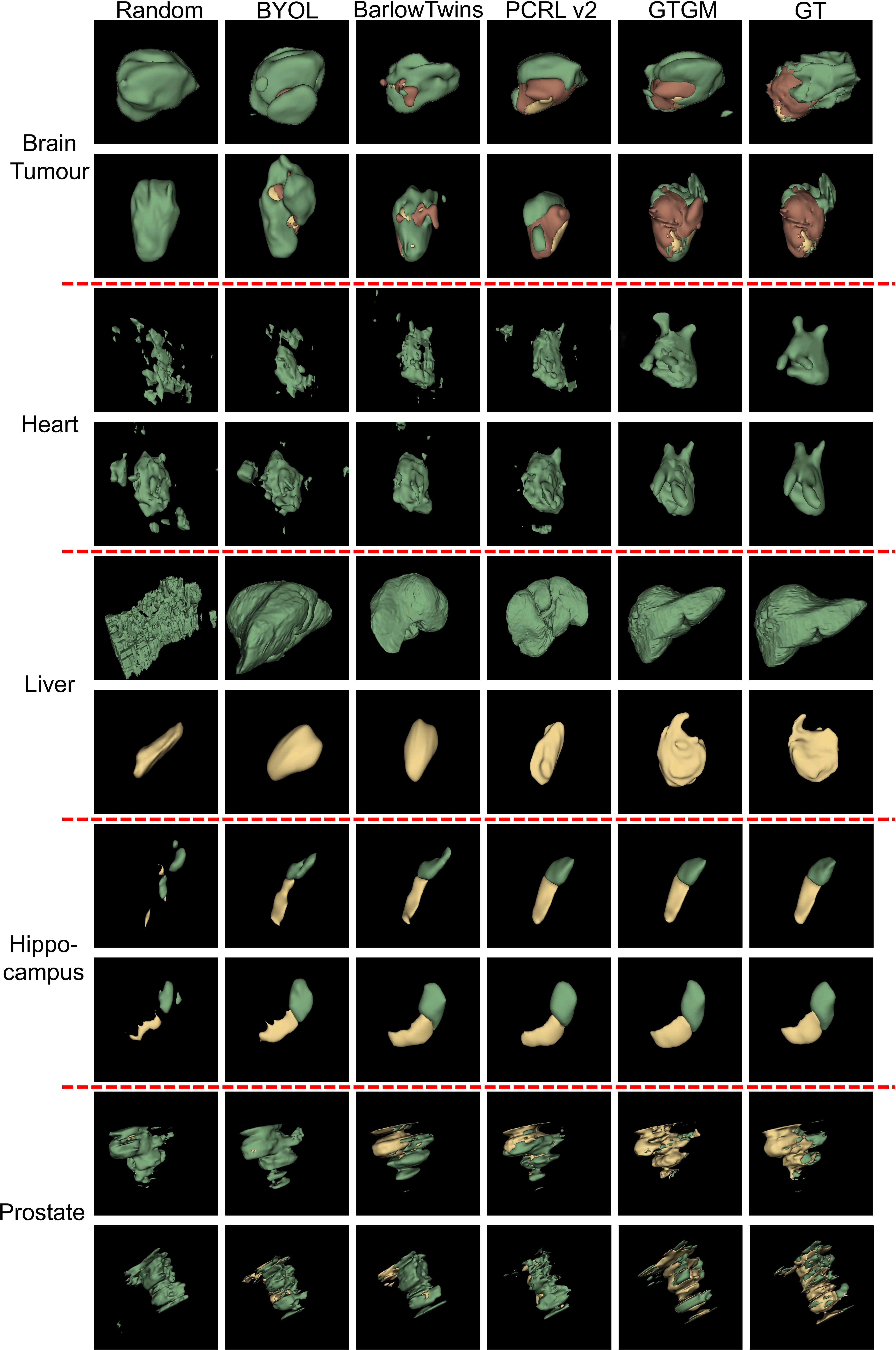}
    \caption{Visualization results of the first 5 tasks of MSD. `GT' indicates ground truth.}
    \label{fig:my_label1}
\end{figure}

\begin{figure}[htb]
    \centering
    \includegraphics[width=\linewidth]{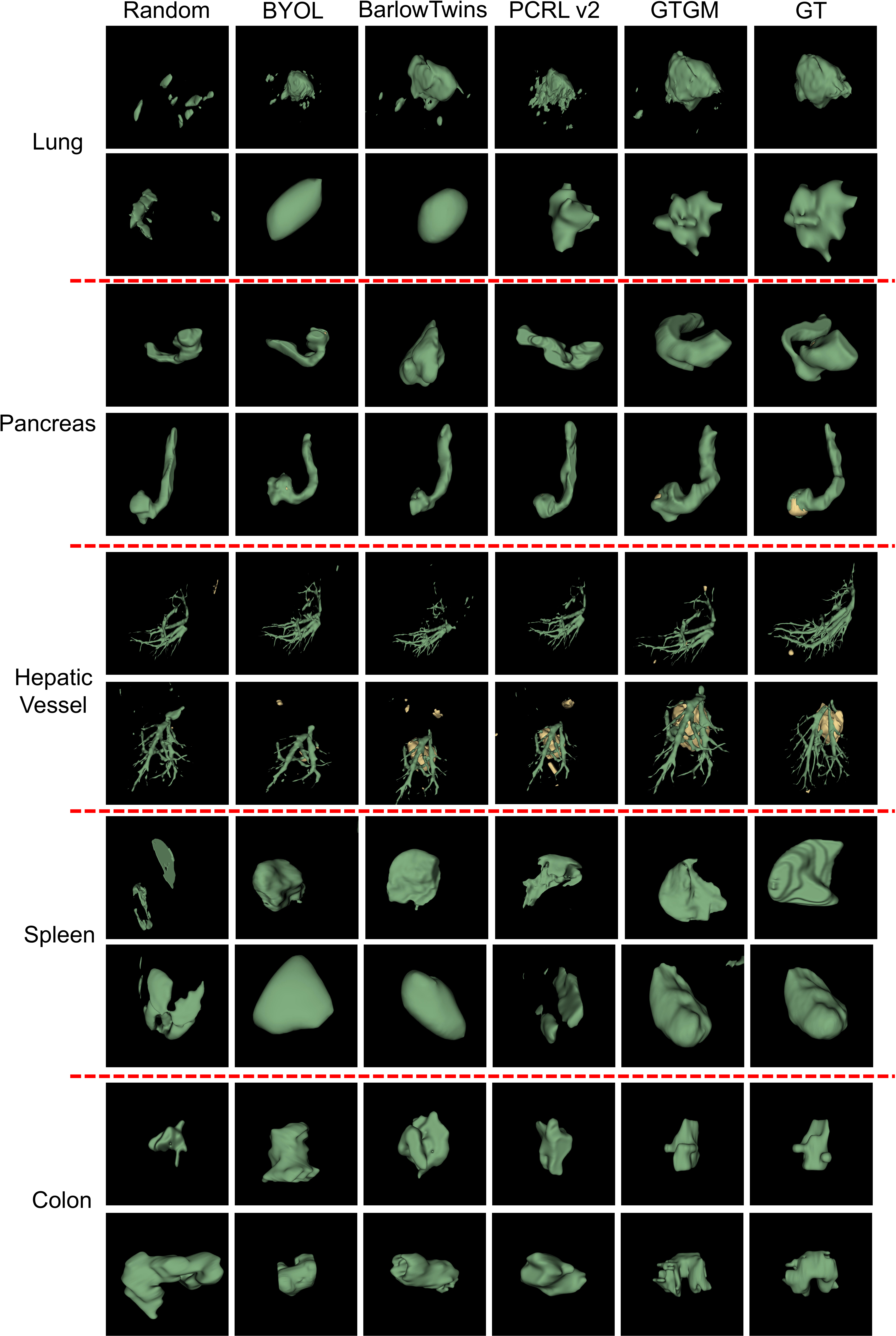}
    \caption{Visualization results of the last 5 tasks of MSD}
    \label{fig:my_label2}
\end{figure}

We present the visualization results of the segmentation task on the MSD dataset \cite{antonelli2022medical}, as shown in Figure \ref{fig:my_label1}, \ref{fig:my_label2}. Our approach demonstrates superior capability in capturing detailed anatomical structures (tumors) compared to other pretrained methods. Specifically, our method exhibits noticeable qualitative improvement, particularly for challenging instance segmentation tasks involving intricate structures such as Hepatic Vessels. The enhanced visual results highlight the efficacy of our approach in accurately delineating the morphological characteristics of organs (tumors) within the medical imaging context.

\subsection{Caption Generate}
\begin{figure}[htb]
    \centering
    \includegraphics[width = 0.9\linewidth]{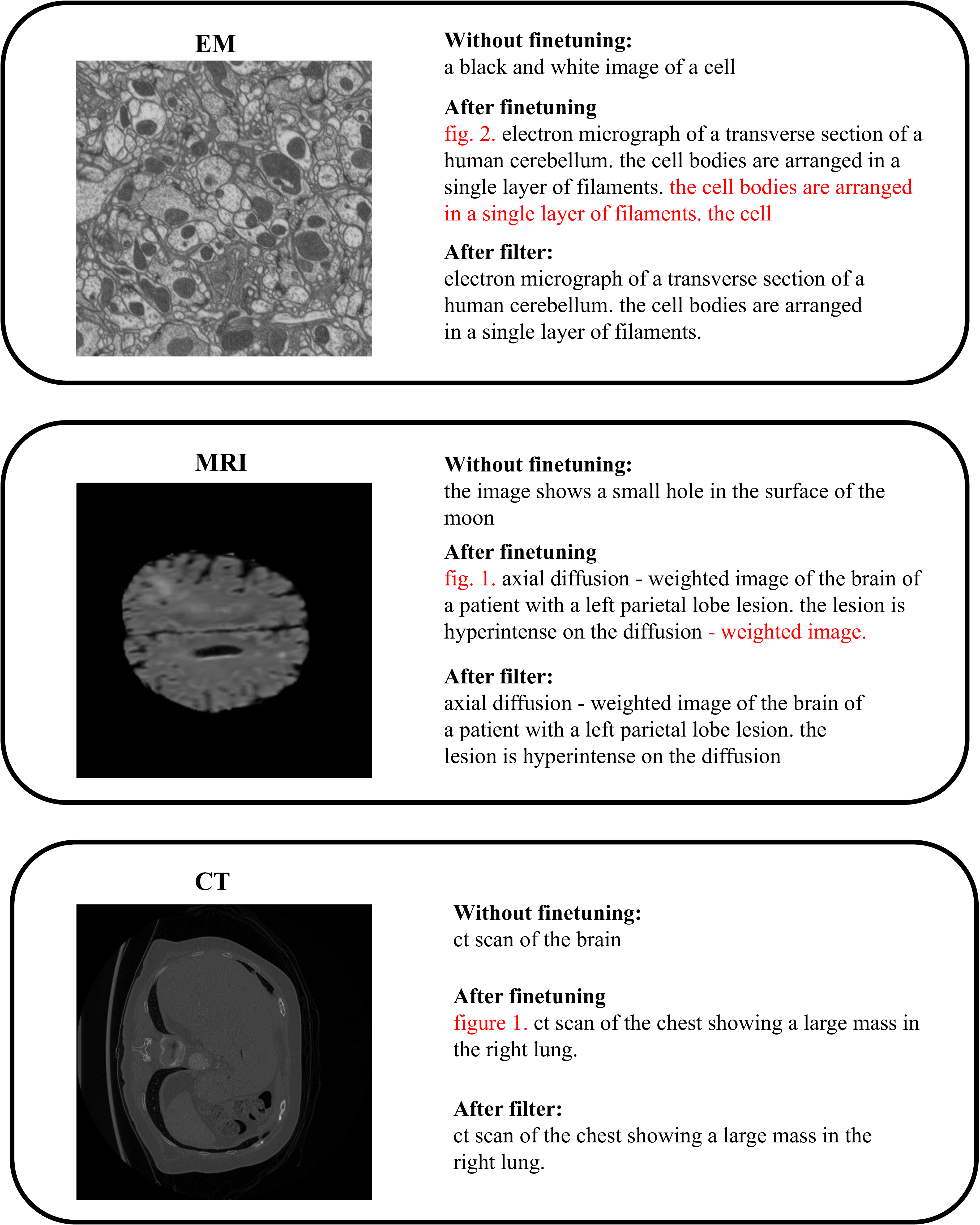}
    \caption{Generated medical description}
    \label{fig:caption}
\end{figure}

We present the generated medical text descriptions in Figure \ref{fig:caption}. The descriptions generated by the untuned large-scale language models (e.g., BLIP~\cite{li2022blip}) exhibit limited information, often resembling natural image descriptions, and contain numerous errors. For example, despite correctly identifying the CT image, there are instances where lung slices are incorrectly labeled as brain slices, introducing misleading information.

However, with our proposed fine-tuning approach, the generated text contains a substantial amount of useful information. Furthermore, utilizing our introduced filter, redundant and repetitive information in the text is eliminated (as indicated by the {\color[HTML]{FF0000} red} text in the figure), thereby enhancing the information density and descriptive accuracy of the generated text.
\clearpage

\section{Pseudo Code}

\begin{figure}[H]
    \centering
    \includegraphics[width = \linewidth]{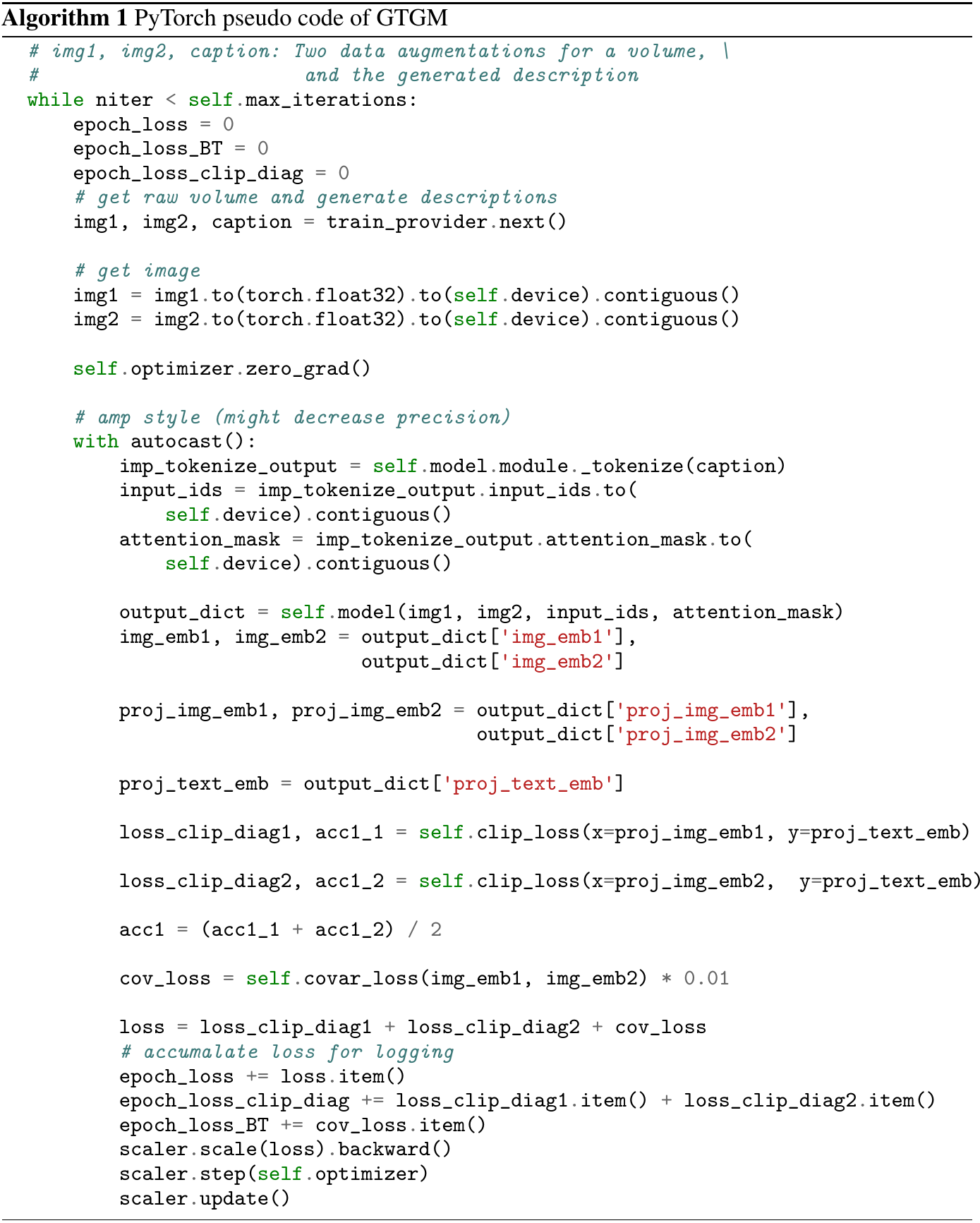}
    \label{alg:GTGM}
\end{figure}

The core code for our pretraining process is outlined in Algorithm \ref{alg:GTGM}.
\vspace{-0.2cm}
\vspace{-0.3cm}


\clearpage
\newpage
{\small
\bibliographystyle{plainnat}
\bibliography{main.bib}
}
\end{document}